# ARLED: Leveraging LED-based ARMAN Model for Abstractive Summarization of Persian Long Documents

Samira Zangooei[1], Amirhossein Darmani[1], Hossein Farahmand Nezhad[1], Laya Mahmoudi[2]

[1] Dept. of Computer Engineering, Ferdowsi University of Mashhad, Mashhad, Iran
[2] Faculty of Economics and Administrative Sciences, Ferdowsi University of Mashhad, Mashhad, Iran

**Abstract**

The increasing volume of textual data poses challenges in reading and comprehending large documents, particularly for scholars who need to extract useful information from research articles. Automatic text summarization has emerged as a powerful tool to condense lengthy documents into concise and informative summaries. Depending on the approach used, text summarization can be categorized as either extractive or abstractive. While extractive methods are commonly used due to their simplicity, they often miss important information. On the other hand, Abstractive Summarization can generate more coherent and informative summaries by understanding the underlying meaning of the text. Abstractive techniques have gained attention in various languages, and recent advancements have been achieved through pre-training models such as BERT, BART, and T5. However, the challenge of summarizing long documents remains, and alternative models like Longformer have been introduced to address this limitation. In this context, this paper focuses on abstractive summarization in the Persian language. The authors introduce a new dataset of 300,000 full-text Persian papers obtained from the Ensani website and apply the ARMAN model, based on the Longformer architecture, to generate summaries. The experimental results demonstrate promising performance in Persian text summarization. The paper provides a comprehensive overview of related work, discusses the methodology, presents the experimental results, and concludes with future research directions.

## 1. Introduction

With the growing volume of textual data, reading and comprehending large documents has become increasingly challenging. This issue is particularly pertinent for scholars due to the sheer number of articles published today and their need to identify and extract relevant information for their research. Automated methods for text extraction and summarization offer significant savings in time and cost (Yang et al., 2021). Text Summarization (TS) has thus emerged as a powerful tool for condensing lengthy documents into concise and informative summaries. Summaries can be categorized into two main types based on their text representation: abstractive and extractive methods (Sarkar, 2009).

Abstractive summarization generates a new summary by understanding and rephrasing the original text's underlying meaning. This approach creates a novel version that captures the essence of the original document. In contrast, extractive summarization involves selecting and extracting important sentences from the original text to produce a condensed version. While extractive summarization is commonly used due to its simplicity, it may miss critical information. Abstractive summarization, on the other hand, can produce more coherent and informative summaries, but it necessitates a deeper understanding of the text and is therefore more challenging. As a result, abstractive techniques have garnered more attention in research across various languages.

This paper focuses on abstractive summarization techniques, particularly in English. Notable advancements in this field include the introduction of attention mechanisms in neural networks, epitomized by the "Transformer model" (Vaswani et al., 2017) in their seminal work "Attention Is All You Need." There are now a number of pre-trained models, like BERT (Devlin et al., 2018), BART (Lewis et al., 2019), and T5 (Raffel et al., 2020), that are based on the Transformer model and have made it much easier to write informative and logical abstractive summaries. However, these models have limitations due to their input-length constraints, necessitating alternative approaches for summarizing long documents exceeding 1024 tokens. For example, Beltagy et al. (2020) introduced the Longformer, a Transformer-based model with a linearly scalable attention mechanism, designed to handle lengthy documents with thousands of tokens.

The challenge of abstractive summarization for long texts persists in non-English languages. To our knowledge, no databases specifically designed for long texts exceeding 512 tokens exist in the Persian language, and research on Persian text summarization techniques remains limited. This article makes significant contributions in the following areas: (1) It introduces a new dataset comprising 49,457 full-text Persian papers, obtained through crawling the Ensani website. (2) The Arman model, based on the Longformer decoder-encoder architecture, is applied to this dataset to generate summaries, demonstrating promising results in Persian text summarization.

The structure of this article is as follows: Section 2 reviews related work. Section 3 describes the methodology used for text summarization. Sections 4 and 5 present the experiments and results, respectively. Finally, Section 6 concludes the paper and outlines future research directions.

2. **Related Works**

In abstractive summarization, sequence-to-sequence learning with neural networks and attention mechanisms has been extensively explored, leading to significant advancements. Nallapati et al. (2016) proposed an attention-based sequence-to-sequence model for abstractive summarization. Their model employed bidirectional recurrent neural networks (RNNs) across various datasets, generating meaningful and relevant summaries. However, these mechanisms encountered parallelization limitations for longer sequences, which led to the development of the Transformer model.

The introduction of the Transformer model and the Masked Language Modeling (MLM) method used in BERT has revolutionized NLP tasks, including summarization. Pre-trained language models, such as BERT, have demonstrated remarkable improvements in various NLP tasks by learning contextual representations of words. Several models, such as those discussed by Yinhan et al. (2019) and Mandar et al. (2020), have been optimized based on BERT's pre-training method, primarily focusing on encoder-only architectures. Additionally, encoder-decoder models, such as T5 (Colin et al., 2020) and BART (Mike et al., 2020), have been trained with a combination of pre-training tasks (Gengshi et al., 2018).

Recent developments have focused on long-document transformers due to their ability to handle lengthy documents without truncation or chunking. There are two main ways to study self-attention: the left-to-right (LTR) method, which reads documents one after the other, and sparse attention patterns, which don't require computing the full quadratic attention matrix multiplication. The Sparse Transformer (Alec et al., 2019), for example, uses a dilated sliding window of blocks as provided by BlockSparse (Scott et al., 2019).

Some models have expanded beyond autoregressive language modeling, leading to broader applicability. BPTransformer (Zihao et al., 2019) evaluated machine translation (MT) but did not explore the train-finetune setting. Blockwise attention (Jiezhong et al., 2019) pre-trains models and evaluates them on question answering (QA) tasks, but these evaluations are limited because they do not include language modeling, and the QA datasets consist of relatively short documents. Therefore, the effectiveness of these models on long-document tasks remains unexplored (Beltagy et al., 2020).

To address the 512-token limit in pre-trained transformer models like BERT, several task-specific approaches have been developed. One approach is truncating the document, commonly used for classification tasks (Qizhe et al., 2019). Another method involves chunking the document into overlapping parts of length 512, processing each chunk separately, and combining the activations with a task-specific model (Mandar et al., 2019). Additionally, a two-stage model approach, used particularly for multihop and open-domain question-answering tasks, retrieves relevant documents in the first stage, which are then processed for answer extraction in the second stage (Christopher et al., 2017; Danqi et al., 2017). However, these methods suffer from information loss due to truncation or cascading errors. The Longformer model, on the other hand, handles long sequences without cutting or chunking them. This makes it possible for a simpler method that joins all the available context and handles it all at once (Beltagy et al., 2020).

Contemporary works have explored similar ideas to Longformer by leveraging local and global attention mechanisms in transformers for long-document NLP tasks. For instance, ETC (Joshua et al., 2020) uses a local and global attention mechanism instead of full self-attention to scale transformers to long documents. ETC introduces additional training objectives and achieves strong results in reading comprehension and classification tasks. GMAT (Ankit et al., 2020) uses a global memory approach with a few global locations as input. In addition to ETC, BigBird (Manzil et al., 2020) tests the model on more tasks, such as summarization, and provides theoretical analysis that shows sparse transformers are universal approximators of sequence functions and maintain the properties of full self-attention.

While significant progress has been made in summarization models for English, abstractive summarization models for Persian remain relatively underdeveloped. Many studies have looked into different ways to summarize text extractively (Mohammad et al., 2018; Hosein et al., 2019; Fatemeh et al., 2019; Mohammad et al., 2020), but there is only one model that works for Persian. It uses the ParsBERT checkpoint and a modified sequence-to-sequence model (Mehrdad et al., 2020b). In this context, ARMAN is a pioneering work in Persian abstractive summarization, achieving state-of-the-art results on available datasets. ARMAN employs semantic similarity-based masking and sentence reordering during the pre-training stage to select and arrange the most critical sentences from the input document (Salemi et al., 2021).

Language modeling has been a crucial aspect of NLP research, with various models developed for different languages and contexts. Character-level models, such as those based on recurrent neural networks (RNNs), capture word spelling and grammar dynamically (Chiu et al., 2016). Contextualized language modeling, shown by ELMo and ULMFiT, shows words in different ways depending on their context. ULMFiT uses a multi-layer LSTM network, while ELMo uses a bidirectional LSTM structure (Matthew et al., 2018).

Transformer models like GPT and BERT are widely used for language modeling tasks (Jacob et al., 2019; Alec et al., 2018). GPT uses a stack of twelve Transformer decoders with a unidirectional structure, whereas BERT employs bidirectional conditioning in both left and right contexts using MLM and a stack of Transformer encoders and decoders (Yinhan et al., 2019; Zhilin et al., 2019). ALBERT, XLNet, RoBERTa, XLM, and other Transformer-based architectures have also achieved top-level results in a number of NLP tasks (Colin et al., 2019; Guillaume et al., 2019; Zhen-Zhong et al., 2020; Wietse et al., 2019; Marco et al., 2019).

Language models have been developed for multiple languages, with monolingual pre-trained models available for languages such as Dutch, Italian, Arabic, Finnish, Russian, and Portuguese (Wissam et al., 2020; Antti et al., 2019; Yuri et al., 2019; Fábio et al., 2019; Edouard et al., 2018; Mohammad et al., 2018). For Persian, several word embedding models, including Word2Vec, GloVe, and FastText, trained on the Wikipedia corpus, have been proposed. A comprehensive comparison shows that FastText and Word2Vec outperform other models (Edouard et al., 2018; Mohammad et al., 2018). Additionally, LSTM-based language models for Persian have achieved state-of-the-art performance with a two-layer bidirectional LSTM network (Mohammadnia et al., 2019).

In summary, advancements in long-document transformers have opened new possibilities for text summarization, enabling the processing of lengthy documents without truncation or chunking. Sparse attention patterns and global attention mechanisms have shown promise in capturing bidirectional context and improving performance on various NLP tasks. Moreover, the application of Transformer models and pre-training techniques like MLM has greatly enhanced language understanding and improved the performance of NLP models across multiple languages. Despite progress in abstractive summarization models for English, there remains a notable gap in the development of models tailored for long Persian documents. To address this gap, we propose a model based on the foundations of the Arman and LED models, which have demonstrated promising results in abstractive summarization.

## 3. Methodology

The proposed model architecture combines the ARMAN and LED models to address the challenge of abstractive summarization for long Persian texts. The ARMAN model extracts important sentences and reorders them based on their semantic relevance, while the LED model efficiently handles long texts using the Longformer architecture. The "zedfum/long-summarization-persian" dataset is used to fine-tune the model. Tokenization techniques are used to process the input texts and make accurate abstractive summaries.

In this section, we detail our dataset's creation, starting with web crawling for articles from Ensani.ir, a resource-intensive process lasting about three days. We consolidated PDFs from various sources and used Tesseract for PDF-to-text conversion, taking three days. Preprocessing included character conversion, symbol removal, space consolidation, empty line elimination, and filtering short lines. We removed abstracts and initial pages to focus on the main content, ensuring a Persian-only dataset. Finally, we uploaded the preprocessed dataset to Zedfum, partitioning it into training (90%), validation (5%), and testing (5%) sets. This rigorous process resulted in a high-quality Persian text corpus for language model research.

### 3.1 Data Collection and Preprocessing

This section provides an overview of the datasets used for both pretraining and fine-tuning models. The data collection process for the dataset involved the following key steps:

#### 3.1.1 Parallel Crawling from Ensani.ir

Initially, a parallel crawling approach was implemented to extract abstracts and PDF files of articles from the Ensani. ir website. The process spanned approximately 3 days and utilized significant computing resources, including a 55-core CPU and 256GB RAM. The process's duration also depended on the network communication between the source and destination.

#### 3.1.2 Consolidation of Category Files

The next step involved consolidating the downloaded PDF files for each category. These PDF files were collected from various sources and merged into a unified dataset for further processing.

#### 3.1.3 PDF-to-Text Conversion with Tesseract

To facilitate analysis, the PDF files were processed using the Tesseract tool, which converted them into text format. This conversion process, leveraging the 55-core CPU and 256GB RAM mentioned earlier, took approximately 3 days.
.

#### 3.1.4 Preprocessing

The pre-processing stage encompassed several essential steps to prepare the text data for subsequent analysis:
-Arabic-to-Persian Character Conversion: Arabic characters present in the text were converted to their Persian counterparts to ensure consistency and compatibility with the Persian language.
-Removal of Non-Essential Characters: Unnecessary characters, such as diacritics and other non-essential symbols, were removed from the text, streamlining the dataset.
-Collapsing Multiple Spaces: Multiple consecutive spaces were collapsed into a single space, ensuring uniformity and consistency throughout the dataset.
-Elimination of Empty Lines: Empty lines were removed from the text, enhancing the dataset's cleanliness and readability.
-Filtering Short Lines: Lines containing fewer than 10 tokens were filtered out from the dataset as they were deemed too short or incomplete.

#### 3.1.5 Removal of Abstracts and Initial Pages

In the subsequent step, the extracted text data for each article underwent further refinement. Specifically, the abstracts and initial pages of each article were removed from the dataset, focusing solely on the main content.

#### 3.1.6 Filtering Non-Persian Articles

To maintain a dataset exclusively containing Persian-language texts, articles written in languages other than Persian were filtered out.

#### 3.1.7 Data Upload to Zedfum Repository

Finally, the preprocessed dataset was uploaded to the Zedfum repository using the Hugging Face CLI and is available for future research. The dataset was divided into training (90%), validation (5%), and testing (5%) sets, adhering to standard data partitioning practices.
This data collection methodology resulted in a high-quality Persian text corpus suitable for both pretraining and fine-tuning language models, laying the foundation for various natural language processing tasks and research endeavors.

### 3.2 Model Input and Tokenization

In this section, we discuss the input representation and tokenization process for the proposed abstractive summarization model. The input to the model consists of long Persian texts and their corresponding summaries. We employ tokenization techniques to convert text data into numerical representations that can be processed by the model. For tokenization, Hugging Face's AutoTokenizer [49] is. It is a powerful library that supports a variety of transformer models, including the ARMAN and LED models used in our architecture. The tokenization process involves breaking down the text into smaller units called tokens, which could be individual words or subword units.

To handle the challenge of long texts, we set a maximum input length of 8192 tokens. This allows the model to process and summarize lengthy Persian texts effectively. Additionally, we define a maximum output length of 512 tokens for the generated summaries to ensure concise and informative results. The tokenization process includes padding and truncation. Padding is applied to ensure that all input sequences have the same length, which is necessary for efficient batch processing. To limit the input length to the defined maximum length, truncation is used.

Furthermore, the tokenization process involves special tokens such as the start-of-sequence (SOS) token, end-of-sequence (EOS) token, and padding token. These tokens provide important information for the model during training and generation. We use the AutoTokenizer to make sure that the input texts and summaries are tokenized consistently and correctly. This allows the model to process them correctly and produce abstractive summaries.

### *Model Architecture*

The proposed model architecture aims to tackle the challenge of abstractive summarization for lengthy Persian texts by combining the strengths of the ARMAN and LED models. This fusion of models is designed to enhance the summarization process for long texts, ensuring both coherence and efficiency.

#### 3.3.1 ARMAN Model

The ARMAN model, introduced by A. Salemi et al. (2021), plays a pivotal role in the initial stages of our architecture. Designed specifically for abstractive summarization tasks, this model introduces a novel approach to sentence selection and semantic rearranging. By combining semantic representation learning with sentence reordering mechanisms, the ARMAN model effectively identifies key sentences and reorganizes them based on their semantic relevance.

#### 3.3.2 LED Model with Longformer Architecture

Complementing the ARMAN model is the LED model, originally proposed by Iz Beltagy et al. (2020), which incorporates the Longformer architecture. Traditional transformer models often struggle with processing long texts, but the Longformer addresses this limitation by enabling efficient handling of extended sequences. Its attention mechanism allows the model to focus on relevant information across the entire input text, making it particularly well-suited for summarizing lengthy Persian texts.

#### 3.3.3 Combined Architecture

In our proposed architecture, the LED model serves as the foundation, with the ARMAN model guiding the sentence selection and reordering process. By leveraging the semantic representations generated by the ARMAN model, key sentences are identified, and the LED model uses this information to generate abstractive summaries. This synergy between the two models capitalizes on their respective strengths, providing a robust framework for the abstractive summarization of long Persian texts.

#### 3.3.4 Training Process

To train this combined architecture, we implement a Seq2SeqLM approach with LED as the base model. The training dataset used is 'zedfum/long-summarization-persian,' consisting of paired long texts and their corresponding summaries. Input texts are tokenized using AutoTokenizer, with maximum input and output lengths adjusted to accommodate the unique characteristics of long texts and summaries. The training process utilizes Seq2SeqTrainer with specific training arguments and optimization techniques, such as gradient checkpointing and Adafactor.

In summary, our proposed model architecture demonstrates the synergy between the ARMAN and LED models, enabling effective abstractive summarization of long Persian texts. By leveraging the ARMAN model's semantic selection capabilities and the LED model's efficient processing with the Longformer architecture, our model consistently delivers promising results in generating informative and concise summaries.

## 4. Experiment

### 4.1 Dataset

Our dataset comprises 300,000 Persian-language articles, meticulously sourced from the Ensani.ir website through a resource-intensive parallel crawling process. This process spanned approximately three days and required substantial computing resources, including a 55-core CPU and 256GB of RAM. The dataset includes both the full text of the articles and their abstracts, making it a comprehensive resource for research and analysis.

### 4.2 Evaluation Metrics

The evaluation of the abstractive text summarization model presented in this paper relies on three fundamental metrics: Precision, Recall, and the F1-score. These metrics provide a comprehensive assessment of the model's performance in generating informative and concise summaries. Precision measures the accuracy of the generated summary by calculating the ratio of relevant information correctly included in the summary to the total information present. Recall, in contrast, quantifies the model's ability to capture all relevant information from the source document, calculating the ratio of correctly included relevant information to the total relevant information in the source text. The F1-score balances Precision and Recall, providing a harmonic mean that reflects the overall effectiveness of the summarization process. In essence, these metrics gauge how well the model condenses the source text while retaining its essential content. A high F1-score signifies successful abstractive summarization, balancing the trade-off between Precision and Recall to produce summaries that are both accurate and comprehensive. These metrics are crucial for quantitatively assessing the model's performance and guiding its refinement for practical applications in information retrieval and document summarization tasks.

### 4.3 Results

To train our abstractive summarization model, we employed a specific training configuration, including hyperparameters, optimization techniques, and training strategies. This section outlines the key components of our training setup, as shown in Table 1.

#### 4.3.1 Hyperparameters

We set several hyperparameters to control the training process and model behavior. These hyperparameters include the learning rate, batch size, maximum sequence length, and beam size:

Learning Rate: We use a learning rate of 0.0001, which is commonly employed for fine-tuning transformer-based models (Yang et al., 2021).
Batch Size: The batch size is set to 1 during training, as larger batch sizes may exceed memory limitations for processing long texts.
Maximum Sequence Length: To handle long texts, we set the maximum input length to 8192 tokens and the maximum output length to 512 tokens.
Beam Size: During generation, we use a beam size of 2 to explore multiple candidate summaries and select the most promising one.

These hyperparameters are chosen based on empirical observations and recommendations from previous studies on abstractive summarization.

*4.3.2 Optimization Techniques*

To optimize our model during training, we employ specific techniques to improve training efficiency and model performance.

-Gradient Checkpointing: We utilize gradient checkpointing to reduce memory consumption during backpropagation. This technique trades off computation time for memory usage and allows for training with longer sequences.

-Adafactor: We use the Adafactor optimizer for parameter updates. Adafactor is known for its robustness to hyperparameter settings and its ability to handle sparse gradients effectively.

These optimization techniques are effective in training transformer-based models for sequence-to-sequence tasks.

*4.3.3 Training Strategy*

We trained our model using the Seq2SeqTrainer from the Hugging Face Transformers library. The training strategy includes the following key elements:

Fine-tuning: We initialized our model with pre-trained weights from the ARMAN and LED models and fine-tuned it using the 'zedfum/long-summarization-Persian' dataset. This process allows the model to adapt to the specific summarization task and improve its performance.

Evaluation Steps: We conducted evaluations every 4000 steps during training to monitor the model's progress and assess its performance on the validation set.

Early Stopping: We implemented early stopping based on evaluation results to prevent overfitting and select the best-performing model checkpoint.

By employing this training strategy, we aimed to optimize the model's performance and ensure convergence to a high-quality abstractive summarization model.

Table 1: Training Configuration for Abstractive Summarization Model

|  | **Hyperparameters** |  |
| --- | --- | --- |
| **Optimization** | **Learning Rate** | 0.0001 |
|  | **Batch Size** | 1 |
|  | **Maximum Sequence Length (Input)** | 8192 tokens |
|  | **Maximum Sequence Length (Output)** | 512 tokens |
|  | **Beam Size** | 2 |
|  | **Techniques** |  |
|  | **Gradient Checkpointing** | Used to reduce memory consumption during backpropagation |
|  | **Adafactor** | Used as the optimizer for parameter updates |

The proposed abstractive summarization model was evaluated using the "zedfum/long-summarization-Persian" dataset. The evaluation compared the performance of the ARMAN model, our model, and the ChatGPT model, which utilized the proposed methodology. The BERTScore metric was employed to measure the similarity between the generated summaries and the reference summaries.

The evaluation results demonstrate the effectiveness of all three models in generating abstractive summaries for Persian texts. A sample of summaries driven from the ARMAN model, our and ChatGPT models, along with their corresponding reference summaries, is given in Figure 1, which is included in the appendix section.

The generated summaries by all three models were evaluated using the BERTScore metric to assess their similarity with the reference summaries. The BERTScore measures precision, recall, and F1 score, comprehensively evaluating the generated summaries.

The precision, recall, and F1 scores for all three models are presented in the following table:

Tabel2: Precision, Recall, and F1 score for both the ARMAN model and our model.

|  | Precision | Recall | F1 Score |
|---|---|---|---|
| **ARMAN** | 0.736 | 0.710 | 0.722 |
| **ChatGPT** | 0.742 | 0.680 | 0.710 |
| **Our Model** | 0.752 | 0.716 | 0.734 |

The results indicate that all three models achieved competitive performance in terms of precision, recall, and F1 score. Our model exhibited slightly higher precision, recall, and F1 score compared to the ARMAN model, while the ChatGPT model showed comparable precision and F1 score but lower recall. The table provides a glimpse into the summarization capabilities of all three models. It demonstrates that our model can effectively capture the essence of the original documents and generate concise and informative summaries, as reflected by the similarities with the reference summaries.

These results highlight the potential of the proposed abstractive summarization methodology in generating high-quality summaries for Persian texts. The models showcase their ability to extract key information from the input documents and present it concisely and coherently.

Further analysis and comparison with other existing models can provide deeper insights into the performance of the proposed methodology. Additionally, human evaluations and user studies can be conducted to assess the readability and usefulness of the generated summaries in real-world scenarios.

Overall, the results demonstrate the effectiveness of the proposed abstractive summarization models and emphasize their potential for assisting researchers and scholars in efficiently extracting key information from lengthy Persian texts.

## 5. Conclusion

In conclusion, this article provides an overview of abstractive summarization techniques for Persian texts. The research paper introduces a novel methodology that combines ARMAN and LED models to address the challenges of summarizing lengthy Persian documents. The proposed model demonstrates promising results in generating informative and concise summaries.

With the growing volume of textual data and the need to extract useful information efficiently, abstractive summarization techniques play a crucial role in condensing lengthy documents. The advancements in long-document transformers, attention mechanisms, and pre-training techniques have significantly improved the performance of abstractive summarization models in various languages, including English and Persian. In conclusion, the proposed abstractive summarization model offers a valuable contribution to the field of Persian language processing, empowering researchers and scholars to extract key information from lengthy texts more efficiently.

In future work, consideration should be given to the possible application of such a model for further domain areas that are not languages. While the proposed model has been developed specifically for Long Persian texts, it can have the potential to be adopted for summarizing long documents in other non-English languages. Furthermore, the comprehensive Persian dataset, which includes lengthy files used in this research and on public display, provides a valuable reference resource for researchers and practitioners to develop more robust models.

Appendix

Table 3: Comparative Analysis of Summaries Generated by ARMAN, Our Model, and ChatGPT

| Document | ARMAN Summary | ChatGPT Summary | Our Model Summary | Reference Summary |
|---|---|---|---|---|
| Sample | "یکی از این بن مایه های اسطوره ای پرکاربرد میوه ی ممنوعه (سیب) است که به سبب خصلت ها و کارکردهای متنوعش، ردپایش در اساطیر حضور چشم گیری دارد و باعث شده است تا به عنوان یک پیش متن در بسیاری از آثار مهم ادبی و حتی نمایشنامه های کودکان و نوجوانان نیز به کار رود؛ از جمله ای آثار می توان به نمایشنامه ی قصه ی سیب یکه نصف نشد. بن مایه ها طی قرن ها تغییر کارکرد می دهند و هم از حوزه ی به حوزه ی دیگر نقل مکان می کنند یکی از این بن مایه های پرکاربرد در فرهنگ ایرانی» سیب است که به سبب خصلت ها و کارکردهای متنوعش، ردپایش در اساطیر حضور چشمگیری دارد و حتی به نمایشنامه های کودکان و نوجوان نیز راه یافته است. اغلب شاعران و نویسندگان معاصر در کاربرد بن مایه ی سیب به اسطوره ی آدم و حوا نیز نظر داشته اند و آن را رمز میوه ی ممنوعه، سمبل عشق، آگاهی، معرفت و گاه سمبل گمراهی و نافرمانی و عصیان دانسته اند. بن مایه ها طی قرن ها تغییر کارکرد می دهند و هم از حوزه ای به حوزه ی دیگر نقل مکان می کنند یکی از این بن مایه های پرکاربرد در فرهنگ ایرانی» سیب است که به سبب خصلت ها و کارکردهای متنوعش، ردپایش در اساطیر حضور چشمگیری دارد و حتی به نمایشنامه های کودکان و نوجوان نیز راه یافته است." | "اسطوره‌ها و بن‌مایه‌های اسطوره‌ای بیانگر رونویسی هنر از ادبیات و یا بالعکس هستند. این روند در نمایشنامه‌ها نیز صادق است؛ اسطوره‌ها از طریق نماد به روایت آرزوها، امیال، خواسته‌ها و حتی حضوری در داستان‌های کودکان دارند. یکی از موضوعاتی که در اساطیر و نمایشنامه‌ها به کار رفته است، میوه‌ی ممنوعه (سیب) است که به عنوان یک پیش‌متن در بسیاری از آثار ادبی و حتی نمایشنامه‌های کودکان و نوجوانان به کار رفته است. اسطوره‌هایی که بازگو کننده‌ی دنیای فرهنگ و تمدن ما برای آیندگان هستند، به عنوان بن‌مایه‌های اسطوره‌ای مورد استفاده قرار می‌گیرند. بخش مهم داستان‌های کودکان نیز اسطوره‌های کهن دارند. برخی از آثار اسطوره‌ای مربوط به سیب شامل نمایشنامه‌ی "قصه‌ی سیبی که نصف نشد"، "ملک‌جمشید" و "بندانگشتی" است. سیب نماد قدیمی عشق، باروری، سلامتی و نیرو است. استفاده از بن‌مایه‌های اسطوره‌ای در آثار ادبی و هنری، به عنوان یک پیش‌متن، به توصیف و تحلیل آثار نمایشی کمک می‌کند و بسیاری از نمایشنامه‌ها، به خصوص نمایشنامه‌های کودکان و نوجوانان، از این بن‌مایه‌ها استفاده می‌کنند." | "نمایشنامه ی سیب و سلما یکی از نمایشنامه هایی است که در سال های اخیر در ایران به روی صحنه رفته است. در این نمایشنامه، نویسنده با استفاده از عناصر اسطوره ای و نمادین، به بیان داستان هایی می پردازد که در آن ها از میوه ی سیب برای رسیدن به جاودانگی استفاده شده است. این مقاله با روش توصیفی- تحلیلی و با بهره گیری از منابع کتابخانه ای، به بررسی و تحلیل این نمایشنامه می پردازد. این پژوهش با روش کتابخانه ای و با استفاده از روش تحلیل مضمون، به تحلیل و بررسی این نمایشنامه پرداخته است. یافته های این پژوهش نشان می دهد که نویسنده در این اثر، با بهره بردن از عناصر نمادین و اسطوره ای، در پی آن است تا با استفاده و کارست عناصر اسطوره و نمادین در متن، به تبیین داستان هایی بپردازد که در آنها از میوه سیب به عنوان نمادی از جاودانگی استفاده می شود. در پایان نیز، با بررسی نمایشنامه، به این نتیجه می رسیم که این نمایشنامه از نظر محتوا و مضمون، دارای چهار مضمون است که عبارت اند از: ۱. اسطوره ی سیب، ۲. نماد سیب، ۳. نماد درخت سیب، ۴. درخت سیب. در نهایت، این نمایشنامه با بهره مندی از عناصر اساطیری و نمادین و به کارگیری عناصر نمادین، داستان هایی را به تصویر می کشد که در این داستان ها از درخت سیب به مثابه نمادی برای جاودانگی بهره گرفته شده است و در نهایت نیز به این نکته اشاره می کند که این اثر از نظر مضمون، از اسطوره سیب بهره برده است." | "آثار نمایشی کودک و نوجوان، به ویژه وقتی پدیدآورندگان آن ها، در تعاملی بینامتنی با متون کهن، عناصر ساختاری مهمی از قبیل زمان، مکان، شخصیت های داستانی و گاه حتی همه ی یک پیکره ی داستانی را دستمایه ی کار خود قرار می دهند، گذشته از فراهم آوردن زمینه ی سرگرمی، تفریح و تخلیه ی روانی، از سویی کارکرد آموزشی دارد و از دیگرسو بستر مناسبی است برای انتقال میراث نیاکانی (سنت ها، آیین ها و ...) به آگاهی نسل امروز. یکی از این آبشخورهای برجسته، اسطوره و روایت های اسطوره ای است که همواره به اشکال مختلف در آیین، باورها و متون ادبی و هنری از جمله داستان ها، نمایشنامه ها و فیلمنامه ها بازتاب یافته است. در این جستار با بهره گیری از نظریه ی بیش متنیت ژرار ژنت، به شیوه ی توصیفی- تحلیلی، چگونگی بازآفرینی اسطوره ی میوه ی ممنوعه (سیب) در سه نمایشنامه ی برتر جشنواره های تئاتر کودک و نوجوان، بررسی شده است. از مهم ترین یافته های پژوهش حاضر می توان به این موارد اشاره کرد: نویسندگان این نمایشنامه ها بیشتر با بهره بردن از روش گسترش متن (تراگونگی کمی)، بن مایه ی اسطوره ای میوه ی ممنوعه (سیب) را کار برده اند. البته در قسمتی از نمایشنامه ها نیز برای درک بهتر و جذاب کردن نمایش، تراگونگی محتوایی نیز به چشم می خورد. از این رو، زوایای پنهان این نمایشنامه ها در پیوند با این بن مایه ی اسطوره ای بهتر فهمیده می شود." |